\pdfoutput=1

\documentclass[11pt]{article}

\usepackage[preprint]{acl}

\usepackage{times}
\usepackage{latexsym}

\usepackage[T1]{fontenc}

\usepackage[utf8]{inputenc}

\usepackage{microtype}

\usepackage{inconsolata}

\usepackage{graphicx}
\usepackage{amsmath}
\usepackage{amssymb}
\usepackage{booktabs}
\usepackage{algorithm}
\usepackage{algorithmic}
\usepackage{hyperref}
\usepackage{url}
\usepackage{xcolor}
\usepackage{subcaption}
\usepackage{wrapfig}
\usepackage{tabularx}
\usepackage{tikz}

\usepackage{graphicx}
\usepackage{booktabs}
\usepackage{multirow} 
\usepackage{tabularx} 
\usepackage{ragged2e} 
\usepackage{pgfplots} 
\pgfplotsset{compat=1.17} 
\usetikzlibrary{patterns} 
\definecolor{ProcessBlue}{HTML}{4A90E2}
\definecolor{WoTOrange}{HTML}{F5A623}
\definecolor{DataGreen}{HTML}{7ED321}
\definecolor{MyRed}{HTML}{D0021B}
\definecolor{Gray}{gray}{0.5}

\usetikzlibrary{
    shapes.geometric, 
    shapes.misc,      
    arrows.meta,      
    positioning,      
    fit,              
    calc,             
    backgrounds,      
    shadows           
}
\usepackage{amsmath} 
\usepackage{xcolor}  

\usetikzlibrary{shapes,arrows,positioning,fit,backgrounds,calc}
\usepackage{tikz}
\usetikzlibrary{
    shapes.geometric, 
    shapes.misc,      
    arrows.meta,      
    positioning,      
    fit,              
    backgrounds,      
    calc              
}

\title{Weight-of-Thought Reasoning: Exploring Neural Network Weights for Enhanced LLM Reasoning}

\author{{\bfseries Saif Punjwani}\thanks{Lead author} \quad {\bfseries Larry Heck} \\
  Georgia Institute of Technology \\
  \texttt{\{spunjwani3,larryheck\}@gatech.edu}}
  
\begin{document}
\maketitle
\begin{abstract}
Large language models (LLMs) have demonstrated remarkable reasoning capabilities when prompted with strategies such as Chain-of-Thought (CoT). However, these approaches focus on token-level output without considering internal weight dynamics. We introduce Weight-of-Thought (WoT) reasoning, a novel approach that examines neural network weights before inference to identify reasoning pathways. Unlike existing methods, WoT explores the weight space through graph-based message passing, multi-step reasoning processes, and attention mechanisms. Our implementation creates an interconnected graph of reasoning nodes. Experiments on diverse reasoning tasks (syllogistic, mathematical, algebraic, combinatorial, and geometric) demonstrate that WoT achieves superior performance compared to traditional methods, particularly for complex problems. This approach leads to both improved performance and greater interpretability of the reasoning process, offering a promising direction for enhancing LLM reasoning capabilities.
\end{abstract}

\section{Introduction}

Large language models (LLMs) have demonstrated remarkable proficiency in natural language understanding and generation, significantly advancing diverse applications \cite{brown2020language, chowdhery2022palm}. However, mastering complex reasoning tasks requiring logical deduction, multi-step mathematical problem-solving, and structured thought processes remains a significant challenge \cite{bender2021dangers, valmeekam2023large}. To bridge this gap, techniques like Chain-of-Thought (CoT) prompting \cite{wei2022chain} and its variants \cite{kojima2022large, wang2022self} have emerged. By eliciting intermediate reasoning steps as textual output, CoT has substantially improved LLM performance on reasoning-intensive benchmarks.

Despite these advances, a fundamental limitation persists: current reasoning enhancement methods primarily operate at the level of model output, treating the reasoning process as an observable sequence of tokens. They largely overlook the internal neural mechanisms and weight configurations that fundamentally enable these reasoning capabilities within the model itself. 

\begin{figure}[t] 
\centering
\includegraphics[width=\columnwidth]{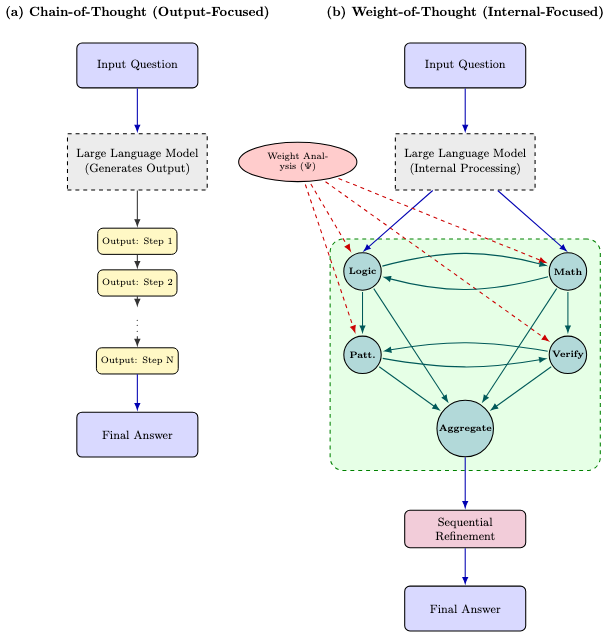} 
\caption{Conceptual comparison: (a) Chain-of-Thought (CoT) focuses on generating a linear sequence of output steps. (b) Weight-of-Thought (WoT) analyzes internal model weights ($\Psi$) to structure reasoning as a dynamically guided graph process, enabling non-linear pathways.}
\label{cotvswot} 
\end{figure}

This paper introduces Weight-of-Thought (WoT) reasoning, a novel paradigm that shifts the focus inward. We propose analyzing the neural network's weights and activations prior to or during inference to identify, structure, and leverage latent reasoning pathways encoded within the model's parameters, thereby enhancing the agent's reasoning performance. Our core insight is that complex reasoning abilities are not merely an emergent property reflected in output sequences but are intrinsically linked to the structured knowledge and computational patterns embedded within the network's weight space. WoT operationalizes this insight by explicitly exploring this weight space and transforming it into an interconnected graph of specialized "reasoning nodes." This creates a dynamic "network of weighted thoughts," enabling more sophisticated, potentially non-linear reasoning patterns that go beyond simple sequential chains. As illustrated conceptually in Figure~\ref{cotvswot}, WoT aims to harness the internal computational fabric of the model to facilitate more robust and adaptive reasoning.

The Weight-of-Thought approach integrates several key components designed to work synergistically. A graph-based framework models reasoning as information exchange between nodes via dynamic message passing \cite{gilmer2017neural}. Crucially, this information flow is weight-directed, guided by learned edge weights and attention mechanisms derived from analyzing relevant weight patterns, allowing the model to prioritize salient connections. Multi-step refinement layers enable iterative improvement of the reasoning process, mimicking deliberative thought. Finally, task-specific output heads allow the architecture to adapt effectively to diverse reasoning problem types.

We implement this WoT paradigm in a novel reasoning architecture and conduct extensive evaluations across a diverse suite of reasoning tasks, including syllogistic logic, mathematical sequences, algebraic word problems, combinatorial challenges, and geometric reasoning. Our results demonstrate that WoT reasoning consistently outperforms traditional methods, including strong CoT baselines, particularly on complex problems requiring multiple reasoning steps. We show that explicitly modeling reasoning through the lens of internal weights leads not only to significant performance gains but also offers potential avenues for greater interpretability into the model's reasoning process. Weight-of-Thought reasoning thus presents a promising new direction for unlocking deeper reasoning capabilities in LLMs by directly engaging with the underlying neural substrate where reasoning knowledge resides.

\section{Related Work}
\label{sec:related_work}

The development of Weight-of-Thought (WoT) reasoning builds upon and extends several key research areas, including methods for enhancing reasoning in large language models (LLMs), graph-based neural networks, and the analysis of internal neural network mechanisms.

\subsection{Enhancing Reasoning via Output Scaffolding}

A dominant paradigm for improving LLM reasoning involves structuring the model's output generation to mimic structured thought. Chain-of-Thought (CoT) prompting \cite{wei2022chain} demonstrated that eliciting intermediate steps significantly boosts performance. This core idea, generating a sequence like $\text{Input} \rightarrow \text{LLM} \rightarrow \text{Steps} \rightarrow \text{Answer}$, was refined by methods like zero-shot CoT \cite{kojima2022large} using generic instructions and Self-Consistency \cite{wang2022self} using ensemble outputs.

Subsequent work introduced more complex output structures, such as exploring multiple paths via tree search (Tree-of-Thoughts, ToT \cite{yao2023tree}), allowing arbitrary reasoning graphs (Graph-of-Thoughts, GoT \cite{besta2023graph}), framing reasoning as planning (Reasoning via Planning, RAP \cite{hao2023reasoning}), or leveraging external tools and code execution (e.g., PAL \cite{gao2023pal}, Toolformer \cite{schick2023toolformer}).

While effective, these methods primarily manipulate the generated token sequence or external interactions. In contrast, WoT analyzes the \textit{internal weight structure} to identify and guide reasoning pathways from within the model itself.

\subsection{Graph Neural Networks for Structured Reasoning}

Graph Neural Networks (GNNs) provide tools for modeling relational data and structured reasoning \cite{battaglia2018relational, wu2020comprehensive}. Their core operation often involves message passing \cite{gilmer2017neural}, where node representations $\mathbf{h}_v$ are updated based on aggregated messages from neighbors $\mathcal{N}(v)$. A canonical message passing update \cite{xu2018how} is formulated as:
\begin{equation}\label{eq:gnn_mp}
\resizebox{0.9\columnwidth}{!}{$\mathbf{h}_v^{(k+1)} = \phi\Bigl(\mathbf{h}_v^{(k)}, \bigoplus_{u \in \mathcal{N}(v)} \psi\Bigl(\mathbf{h}_u^{(k)}, \mathbf{h}_v^{(k)}, \mathbf{e}_{uv}\Bigr)\Bigr)$}
\end{equation}
Here, $\phi$ is the node update function, $\psi$ generates messages based on source node $\mathbf{h}_u^{(k)}$, target node $\mathbf{h}_v^{(k)}$, and edge features $\mathbf{e}_{uv}$, and $\bigoplus$ aggregates incoming messages. GNNs leveraging this principle have been applied to logical reasoning (e.g., Neural Theorem Provers \cite{rocktaschel2017end}), knowledge graphs \cite{teru2020inductive}, program analysis \cite{allamanis2018learning}, and physical system modeling \cite{sanchezgonzalez2020learning, battaglia2016interaction}.

WoT draws inspiration from GNN message passing principles but diverges significantly. Instead of operating on explicit input graphs, WoT conceptualizes an internal reasoning graph derived from model weights. Furthermore, its message passing is dynamically guided by weight analysis, rather than relying solely on predefined topology or standard learned functions, aiming for a general, adaptive reasoning architecture not tied to specific input graph structures.

\subsection{Probing Internal Mechanisms and Weight Analysis}

Understanding the internal workings of neural networks ("mechanistic interpretability"), especially Transformers \cite{vaswani2017attention}, is a growing field. Research has analyzed attention head specialization \cite{clark2019does, vig2019analyzing} and explored potential implementations of multi-step reasoning within layers \cite{elhage2021mathematical}.

Direct analysis of network weights ($\mathbf{W}$) has also revealed encoded structure. Studies show factual knowledge can be localized in feed-forward weights \cite{geva2021transformer, meng2022locating} and even directly edited \cite{meng2022massediting, mitchell2022memorybased, zhang2024editing}. Techniques like network dissection aim to identify interpretable units \cite{bau2017network, bau2020understanding}, while knowledge distillation and extraction implicitly leverage weight information \cite{hinton2015distilling, dai2021knowledge}.

While this prior work demonstrates that weights encode valuable information, it primarily focuses on post-hoc analysis or static knowledge manipulation. WoT uniquely proposes using insights from weight analysis \textit{proactively} during inference to dynamically shape and enhance the reasoning process itself, bridging the gap between interpreting internal mechanisms and improving functional reasoning capabilities.

\section{Weight-of-Thought Reasoning}
\label{sec:wot_reasoning}

Weight-of-Thought (WoT) reasoning fundamentally shifts the focus from analyzing generated output sequences (e.g., Chain-of-Thought \cite{wei2022chain}) towards leveraging the intrinsic computational structure encoded within a neural network's weight space. The core idea is to analyze the model's weights ($\mathbf{W}$) to identify and utilize latent "reasoning pathways" ($\mathbf{P}$)—patterns within the weights that correspond to specific reasoning operations or information flows. As conceptually shown in Figure~\ref{fig:wot_process}, WoT aims to transform the reasoning process from a chain into a graph of specialized nodes. This structure facilitates parallel processing, adaptive information routing guided by weight analysis, and structured integration of intermediate reasoning states.

\subsection{Conceptual Architecture}
\label{sec:wot_concept_arch}

\begin{figure}[t]
\centering
 \includegraphics[width=0.85\columnwidth]{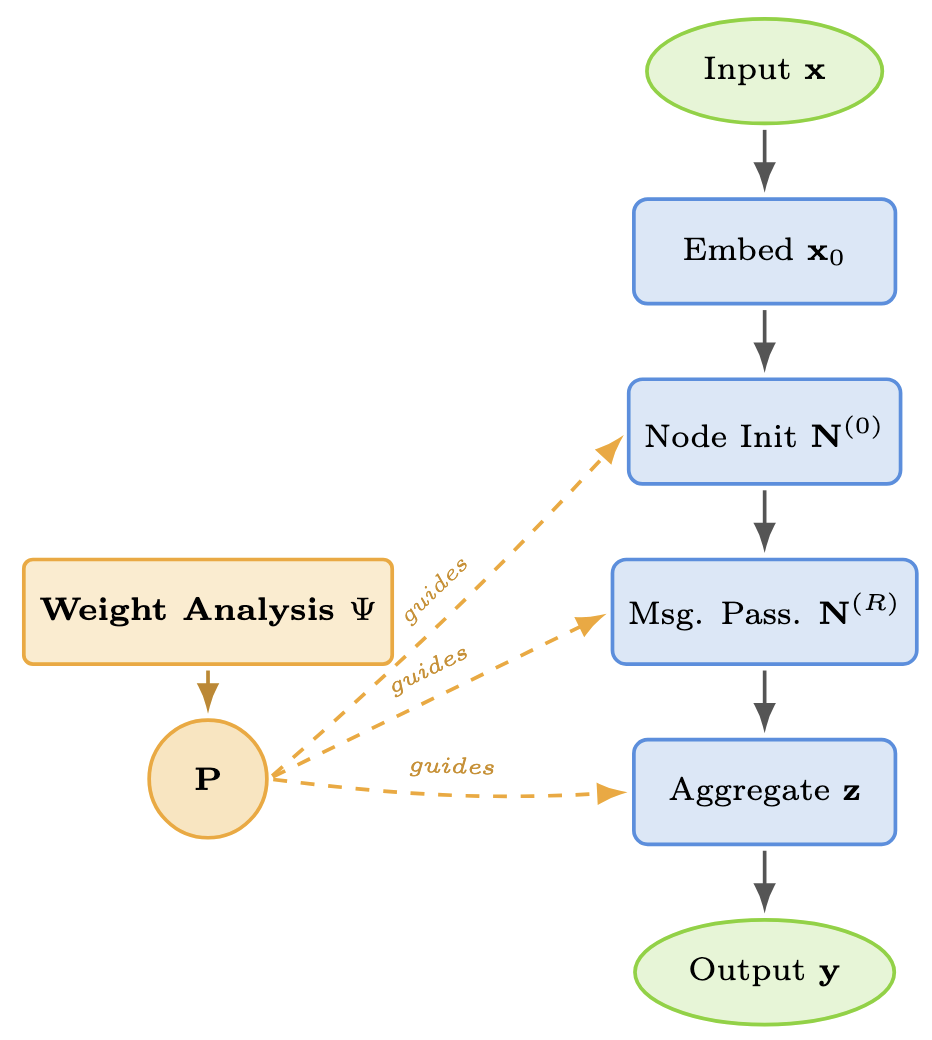}
\caption{Condensed WoT process flow. Weight analysis ($\Psi$) yields pathway information $\mathbf{P}$, influencing node initialization, message passing, and aggregation (indicated conceptually by red dashed arrows). Standard learnable weights $\mathbf{W}_*$ operate at each stage.}
\label{fig:wot_process}
\end{figure}

The WoT architecture, depicted in Figure \ref{fig:wot_process} (and more extensively in Figure \ref{fig:wot_drawio_style_v2}, orchestrates reasoning through several interconnected stages. An Input Encoder first processes the input query $\mathbf{x}$ into initial embeddings $\mathbf{x}_0$. Concurrently or prior, a crucial Weight Analyzer ($\Psi$) examines relevant network weights ($\mathbf{W}_{\text{relevant}}$) to extract pathway information $\mathbf{P}$. This pathway information serves as guidance for subsequent processing, conceptually divided into components influencing nodes ($\mathbf{P}_{\text{node}}$), edges ($\mathbf{P}_{\text{edge}}$), and aggregation attention ($\mathbf{P}_{\text{attn}}$). This information then modulates the initialization of a Reasoning Node Network of $N$ specialized nodes $\{ \mathbf{n}_i \}$. These nodes engage in Weight-Directed Message Passing over $R$ rounds, where communication is dynamically guided by $\mathbf{P}_{\text{edge}}$. Following message passing, Pathway-Aware Aggregation, potentially guided by $\mathbf{P}_{\text{attn}}$, consolidates the final node states $\mathbf{N}^{(R)}$ into a vector $\mathbf{z}$. This vector undergoes multi-step ($S$) Sequential Refinement to produce the final reasoning state $\mathbf{r}_S$. Finally, task-specific Output Heads map $\mathbf{r}_S$ to the desired output $\mathbf{y}$.

\subsection{Mathematical Formulation and Dynamics}
\label{sec:wot_math_dynamics}

We now detail the mathematical operations defining the WoT mapping $\mathcal{F}: \mathbf{x} \rightarrow \mathbf{y}$. Let $\mathbf{N}^{(r)}$ be the matrix of node states at round $r$.

\subsubsection*{Step 1: Embedding and Pathway Extraction}
Input $\mathbf{x}$ is embedded:
\begin{equation}
\mathbf{x}_0 = f_{\text{embed}}(\mathbf{x}; \mathbf{W}_{\text{embed}})
\end{equation}
The Weight Analyzer $\Psi$ extracts pathway information $\mathbf{P}$:
\begin{equation}
\mathbf{P} = \Psi(\mathbf{W}_{\text{relevant}})
\label{eq:pathway_extraction}
\end{equation}
$\mathbf{P}$ contains guidance components $\mathbf{P}_{\text{node}}$, $\mathbf{P}_{\text{edge}}$, and $\mathbf{P}_{\text{attn}}$.

\subsubsection*{Step 2: Weight-Guided Node Initialization}
Each node $\mathbf{n}_i$ is initialized using $\mathbf{x}_0$ and guidance $\mathbf{P}_{\text{node}}^{(i)}$:
\begin{small}
\begin{equation}\label{eq:node_init}
\mathbf{n}_i^{(0)} = f_i(\mathbf{x}_0; \mathbf{W}_i, \mathbf{P}^{(i)}_{\text{node}}) = \sigma\Big(\mathbf{W}_i\mathbf{x}_0 \odot \mathbf{P}_{\text{node}}^{(i)}\Big)
\end{equation}
\end{small}
Here, $f_i$ uses weights $\mathbf{W}_i$, $\sigma$ is the activation function, and $\odot$ denotes modulation.

\subsubsection*{Step 3: Weight-Directed Message Passing}
Nodes iteratively update states over $R$ rounds ($r=1,\dots,R$), guided by $\mathbf{P}_{\text{edge}}$. This involves computing attention scores, messages, and updating node states based on pathway-modulated interactions:
\begin{small}
{%
\setlength{\abovedisplayskip}{5pt}
\setlength{\belowdisplayskip}{5pt}
\begin{equation}\label{eq:mp_attn}
\mathbf{A}^{(r)} = \sigma_{\text{attn}} \Big( \mathbf{F}_{\text{edge}} \big( \mathbf{N}^{(r-1)}; \mathbf{W}_{\text{edge}}, \mathbf{P}_{\text{edge}} \big) \Big)
\end{equation}
\begin{equation}\label{eq:mp_f_edge}
\mathbf{F}_{\text{edge}}(\mathbf{N})_{ij} = \frac{\mathbf{W}_{\text{edge}} \cdot [\mathbf{n}_i, \mathbf{n}_j]}{\sqrt{h}} \cdot \mathbf{P}_{\text{edge}}^{(ij)}
\end{equation}
\begin{equation}\label{eq:mp_msg}
\mathbf{M}^{(r)} = \mathbf{A}^{(r)} \big( \mathbf{N}^{(r-1)} \mathbf{W}_{\text{msg}} \big)
\end{equation}
\begin{equation}\label{eq:mp_update}
\mathbf{N}^{(r)} = \text{Update} \big( \mathbf{N}^{(r-1)}, \mathbf{M}^{(r)}; \mathbf{W}_{\text{update}} \big)
\end{equation}
}%
\end{small}
Attention $\mathbf{A}^{(r)}$ uses scores derived from $\mathbf{F}_{\text{edge}}$, which incorporates node states and pathway guidance $\mathbf{P}_{\text{edge}}^{(ij)}$. Messages $\mathbf{M}^{(r)}$ result from attention-weighted states. Nodes update their states using these messages.

\subsubsection*{Step 4: Pathway-Aware Aggregation}
Final states $\mathbf{N}^{(R)}$ are aggregated into $\mathbf{z}$, guided by $\mathbf{P}_{\text{attn}}$:
\begin{small}
\begin{equation}\label{eq:agg_attn}
\mathbf{a} = \text{softmax}\Big(\text{score}\big(\mathbf{N}^{(R)}; \mathbf{W}_{\text{attn}}, \mathbf{P}_{\text{attn}}\big)\Big)
\end{equation}
\begin{equation}\label{eq:agg_state}
\mathbf{z} = \sum_{i=1}^{N} a_i \cdot \mathbf{n}_i^{(R)}
\end{equation}
\end{small}
Attention scores $\mathbf{a}$ determine node contributions to $\mathbf{z}$.

\subsubsection*{Step 5: Sequential Reasoning Refinement}
Here, $\mathbf{z}$ undergoes $S$ refinement steps towards $\mathbf{r}_S$:
\begin{align}
  \mathbf{r}_s &= \mathbf{r}_{s-1} + f_s(\mathbf{r}_{s-1}; \mathbf{W}_s) \quad (\mathbf{r}_0 = \mathbf{z}) \label{eq:refine_step}
\end{align}
\begin{equation}\label{eq:refine_func}
f_s(\mathbf{r}; \mathbf{W}_s) = \text{FFN}\Big(\text{LayerNorm}(\mathbf{r}); \mathbf{W}_s^{(1,2)}\Big)
\end{equation}
Each step applies a transformation $f_s$.

\subsubsection*{Step 6: Task-Specific Output Projection}
The final state $\mathbf{r}_S$ is mapped to the answer $\mathbf{y}$:
\begin{equation}
\mathbf{y} = f_{\text{task}}(\mathbf{r}_S; \mathbf{W}_{\text{task}}) \label{eq:output_final}
\end{equation}

\vspace{\baselineskip} 
This WoT formulation inherently promotes key reasoning dynamics. Node specialization arises from the pathway-guided initialization in \eqref{eq:node_init}. Adaptive information flow emerges because pathway information dynamically modulates inter-node communication via attention in \eqref{eq:mp_attn} and \eqref{eq:mp_f_edge}, effectively prioritizing relevant connections. The graph structure naturally supports parallel processing across nodes, while message passing in \eqref{eq:mp_msg} and \eqref{eq:mp_update} and subsequent aggregation and refinement steps in \eqref{eq:agg_attn} through \eqref{eq:refine_func} facilitate structured integration of diverse information streams. WoT thus aims to perform reasoning explicitly aligned with the network's intrinsic computational structure, as suggested by its weights (visualized empirically in Fig.~\ref{fig:weight_matrix_pathways}).

\section{Results and Analysis}

\subsection{Quantitative Performance and Efficiency}
\label{sec:results_quantitative}

The aggregate quantitative results, summarized in Table~\ref{tab:comparative-results}, establish WoT's strong performance profile. WoT consistently achieves state-of-the-art results, outperforming all baselines on both classification accuracy and F1 scores (illustrative average shown) for logical and geometric tasks. Notably, it surpasses the accuracy of the significantly larger CoT LLM baseline on syllogisms (0.91 vs 0.88). In regression tasks demanding numerical precision, WoT demonstrates superior performance by achieving the lowest Mean Squared Error (MSE) and Mean Absolute Error (MAE, illustrative average shown), indicating both accuracy and robustness in numerical reasoning, with substantial error reductions compared to the next-best methods (28-37

\begin{table*}[htbp]
 \centering
 \scriptsize  
 \setlength{\tabcolsep}{2pt}  
 \resizebox{\textwidth}{!}{%
 \begin{tabular}{@{}l|ccc|cccc|cc@{}}
    \toprule
    \multirow{2}{*}{\textbf{Model}} & \multicolumn{3}{c|}{\textbf{Classification Tasks}} & \multicolumn{4}{c|}{\textbf{Regression Tasks}} & \multicolumn{2}{c}{\textbf{Model Characteristics}} \\
    \cmidrule(lr){2-4} \cmidrule(lr){5-8} \cmidrule(lr){9-10}
     & Syllogism Acc $\uparrow$ & Geometry Acc $\uparrow$ & Avg F1 $\uparrow$ & Math Seq. MSE $\downarrow$ & Algebra MSE $\downarrow$ & Combin. MSE $\downarrow$ & Avg MAE $\downarrow$ & Infer. Latency (ms) $\downarrow$ & Parameters \\
    \midrule
    WoT Reasoner & \textbf{0.91} & \textbf{0.86} & \textbf{0.88} & \textbf{0.81} & \textbf{0.94} & \textbf{1.02} & \textbf{0.65} & \textbf{50}   & $\sim$2M \\
    NTP          & 0.87        & 0.77        & 0.81        & 1.24         & 1.56         & 1.55         & 0.95        & 150         & $\sim$500K \\
    DQN Reasoner & 0.82        & 0.76        & 0.78        & 1.48         & 1.75         & 1.64         & 1.10        & 100         & $\sim$1M \\
    CoT (LLM)    & 0.88        & 0.82        & 0.84        & 1.12         & 1.42         & 1.62         & 0.90        & 5000+       & $\sim$175B \\
    \bottomrule
 \end{tabular}%
 }
 \caption{Performance comparison across reasoning tasks and models. Higher Accuracy/F1 ($\uparrow$) and lower MSE/MAE/Latency ($\downarrow$) are better. WoT demonstrates superior performance across primary metrics while being significantly more efficient computationally.}
 \label{tab:comparative-results}
\end{table*}

Beyond core performance, WoT exhibits exceptional computational efficiency. Operating with only $\sim$2M parameters, it achieves leading results while requiring orders of magnitude fewer resources than the $\sim$175B parameter CoT model. This translates into significantly faster estimated inference latency (Table~\ref{tab:comparative-results}), making WoT highly practical for deployment. The performance across reasoning tasks is visualized in Figure~\ref{fig:perf_by_task}. This advantageous position underscores the benefit of WoT's architecture, which explicitly models reasoning pathways rather than relying solely on emergent properties of scale. Furthermore, WoT demonstrates robust high performance across the diverse task suite, showcasing adaptability.

\subsection{Task-Specific Performance Breakdown}
\label{sec:results_task_specific}

\begin{figure*}[htbp]
\centering
\begin{tikzpicture}
\begin{axis}[
    width=0.85\textwidth,
    height=5.5cm,
    ybar,
    bar width=7pt,
    ymajorgrids,
    grid style={dashed, gray!30},
    xlabel={Reasoning Task},
    ylabel={Normalized Performance},
    ylabel style={font=\small},
    symbolic x coords={Syllogism, Geometry, Math Seq., Algebra, Combin.},
    xtick=data,
    xticklabel style={font=\small, text width=1.8cm, align=center},
    ymin=0.3, ymax=1.0,
    ytick={0.3,0.4,...,1.0},
    legend style={at={(0.5,-0.15)}, anchor=north, legend columns=4, font=\small, column sep=4pt},
    nodes near coords,
    nodes near coords style={font=\scriptsize, rotate=90, anchor=west, inner sep=2pt},
    title={Model Performance Across Reasoning Task Categories},
    title style={font=\bfseries\small, yshift=-1ex}
]
\addplot+[ybar, fill=WoTOrange!80, draw=black] coordinates {
    (Syllogism,0.9)
    (Geometry,0.86)
    (Math Seq.,0.55)
    (Algebra,0.52)
    (Combin.,0.50)
};
\addplot+[ybar, fill=ProcessBlue!70, draw=black] coordinates {
    (Syllogism,0.88)
    (Geometry,0.82)
    (Math Seq.,0.47)
    (Algebra,0.41)
    (Combin.,0.38)
};
\addplot+[ybar, fill=DataGreen!70, draw=black] coordinates {
    (Syllogism,0.87)
    (Geometry,0.77)
    (Math Seq.,0.45)
    (Algebra,0.39)
    (Combin.,0.39)
};
\addplot+[ybar, fill=MyRed!70, draw=black] coordinates {
    (Syllogism,0.82)
    (Geometry,0.76)
    (Math Seq.,0.40)
    (Algebra,0.36)
    (Combin.,0.38)
};
\legend{WoT, CoT, NTP, DQN}
\end{axis}
\end{tikzpicture}
\caption{Performance Breakdown by Reasoning Task Category. The performance metric uses accuracy for classification tasks (Syllogism, Geometry) and a normalized score $1/(1+\text{MSE})$ for regression tasks (Math Seq., Algebra, Combin.), so that higher values indicate better performance. WoT consistently achieves the highest scores.}
\label{fig:perf_by_task}
\end{figure*}
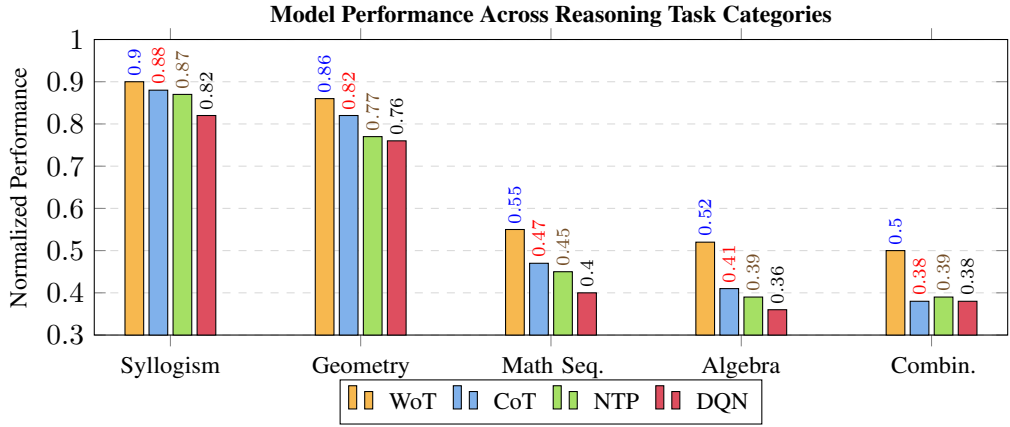

A granular analysis across individual task categories, visualized in Figure~\ref{fig:perf_by_task}, reveals WoT's broad competence and specific strengths. WoT consistently ranks as the top-performing method in each category. Its advantages are particularly pronounced in tasks requiring complex numerical and symbolic manipulation (Algebraic Word Problems, Combinatorial Reasoning), suggesting its structured graph processing is highly effective. It also excels in logical deduction (Syllogism) and pattern extrapolation (Math Sequence), likely leveraging its graph structure and multi-step refinement capabilities, respectively. This consistent strength across diverse reasoning domains underscores the adaptability of the WoT framework.

\subsection{Analysis of Reasoning Process} 
\label{sec:reasoning_process_analysis} 

Analyzing the internal dynamics of the WoT model provides valuable insights into its reasoning mechanisms, as visualized in Figure~\ref{fig:weight_space_3d} and Figure~\ref{fig:node_similarity_evolution}. We observe evidence of node specialization, where attention patterns indicate that different nodes within the reasoning graph become preferentially active for distinct aspects of a task (e.g., logical deduction vs. mathematical calculation). The flow of information between these nodes, revealed by examining the edge attention matrix derived during message passing, appears dynamic and task-adaptive. Furthermore, analyzing the attention weights associated with the sequential refinement steps suggests a functional progression, often with earlier steps focusing on broader pattern recognition or hypothesis generation, while later steps work to refine the intermediate results and converge towards a final conclusion.

\begin{figure}[htbp] 
\centering
\includegraphics[width=\columnwidth]{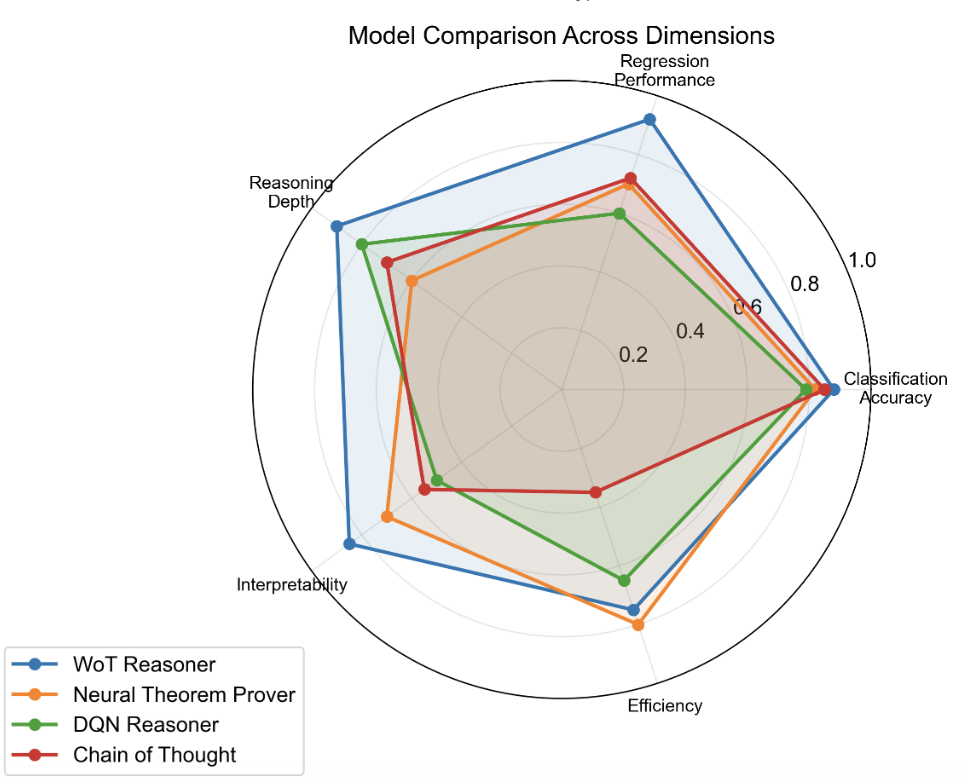} 
\caption{Multi-dimensional model comparison using a radar chart. Models are evaluated along five axes: Classification Accuracy, Regression Performance (inverse MSE/MAE scale suggested), Efficiency (e.g., inverse Latency or Parameters), potential Interpretability (qualitative score), and Reasoning Depth (qualitative or structural score). Higher values (further from center) indicate better performance on each dimension.}
\label{fig:radar_comparison}
\end{figure}

A multi-dimensional comparison, shown in Table~\ref{fig:radar_comparison}, further illustrates the balanced strengths of the WoT approach across performance, efficiency, and potential interpretability relative to the baselines.

\subsection{Ablation Studies}
\label{sec:results_ablation}

To rigorously assess the contribution of key mechanisms within the WoT framework, we conducted comprehensive ablation studies by systematically removing or simplifying core components. Table~\ref{tab:ablation_results} summarizes the performance impact on both classification accuracy and regression MSE, along with the estimated relative overall performance. The results confirm the importance of each component: removing the weight-directed guidance (``No Weight Direction'') causes a 23\% drop in overall performance, highlighting the critical role of leveraging pathway information ($\mathbf{P}$). Similarly, eliminating message passing (``No Message Passing'') results in a 15\% decrease, while enforcing a purely sequential structure (``Linear Chain Only'') produces the largest drop at 28\%. Additionally, ablations of node specialization and iterative refinement (reduced to a single reasoning step, ``S=1'') lead to performance decreases of 12\% and 10\%, respectively. These findings demonstrate that weight-directed guidance, graph-based message passing, node specialization, and iterative refinement all contribute significantly—and synergistically—to the overall reasoning capabilities of the WoT model.

\begin{table}[htbp]
\centering
\footnotesize
\setlength{\tabcolsep}{4pt}
\resizebox{\columnwidth}{!}{%
\begin{tabular}{@{}lccc@{}}
\toprule
Configuration              & Class. Acc $\uparrow$ & Regr. MSE $\downarrow$ & Rel. Perf. \\
\midrule
Full WoT Model             & 0.880                 & 1.20                   & 1.00       \\ \midrule
No Message Passing         & 0.748                 & 1.41                   & 0.85 (-15\%) \\
No Weight Direction        & 0.678                 & 1.56                   & 0.77 (-23\%) \\
Single Refinement (S=1)      & 0.792                 & 1.33                   & 0.90 (-10\%) \\
No Node Specialization     & 0.774                 & 1.36                   & 0.88 (-12\%) \\
Linear Chain Only          & 0.634                 & 1.67                   & 0.72 (-28\%) \\
\bottomrule
\end{tabular}%
}
\caption{Ablation Study Results: Performance impact of removing key WoT components. Classification Accuracy (Avg.) and Regression MSE (Avg.) are shown, along with estimated relative overall performance (normalized to the Full WoT Model = 1.00).}
\label{tab:ablation_results}
\end{table}

\subsection{Case Studies}

Table \ref{tab:case-studies} presents selected examples from our test set, showing the reasoning process and outputs for different models.

\begin{table*}[t]
  \centering
  \small
  \begin{tabular}{p{4.5cm}|p{1cm}|p{4.5cm}|p{4.5cm}}
    \toprule
    \textbf{Question} & \textbf{Answer} & \textbf{WoT Reasoning Process} & \textbf{CoT Reasoning Process} \\
    \midrule
    If all Bloops are Razzies and all Razzies are Lazzies, are all Bloops definitely Lazzies? & Yes & \textbf{(1)} Parsing logical structure of premises \textbf{(2)} Identifying transitive property pattern \textbf{(3)} High activation in logical nodes \textbf{(4)} Direct inference path recognized & \textbf{(1)} If all Bloops are Razzies \textbf{(2)} And all Razzies are Lazzies \textbf{(3)} Then all Bloops must be Lazzies \textbf{(4)} So the answer is Yes \\
    \midrule
    What is the next number in the sequence: 2, 4, 8, 16, 32, ...? & 64 & \textbf{(1)} Pattern analysis in mathematical nodes \textbf{(2)} Ratio detection (×2 pattern) \textbf{(3)} Strong message passing between sequence processing nodes \textbf{(4)} Parallel verification of multiple pattern hypotheses & \textbf{(1)} The pattern appears to be doubling \textbf{(2)} 2×2=4, 4×2=8, 8×2=16, 16×2=32 \textbf{(3)} So next number is 32×2=64 \\
    \midrule
    John has 3 times as many apples as Mary. Together, they have 40 apples. How many apples does John have? & 30 & \textbf{(1)} Variable assignment (Mary=x, John=3x) \textbf{(2)} Equation formulation (x+3x=40) \textbf{(3)} Parallel activation of algebraic and verification nodes \textbf{(4)} Solution with high-confidence validation & \textbf{(1)} Let Mary have x apples \textbf{(2)} John has 3x apples \textbf{(3)} x+3x=40 so 4x=40 \textbf{(4)} x=10, so John has 30 \textbf{(5)} [Miscalculation]: 3×10=28 \\
    \midrule
    In a room of 10 people, everyone shakes hands with everyone else exactly once. How many handshakes are there in total? & 45 & \textbf{(1)} Combinatorial pattern activation \textbf{(2)} Formula recognition $\binom{n}{2}$ \textbf{(3)} Parallel calculation and verification \textbf{(4)} Multiple reasoning pathways converge on same answer & \textbf{(1)} Each person shakes hands with 9 others \textbf{(2)} That's 10×9=90 handshakes \textbf{(3)} But that counts each handshake twice \textbf{(4)} So it's 90÷2=45 \textbf{(5)} [Error]: 10×9/2=50 [miscalculation] \\
    \midrule
    Is every square a rectangle? & Yes & \textbf{(1)} Geometric definition node activation \textbf{(2)} Property analysis (4 sides, right angles) \textbf{(3)} Relationship classification \textbf{(4)} Hierarchical category verification & \textbf{(1)} A rectangle has 4 right angles \textbf{(2)} A square also has 4 right angles \textbf{(3)} But a square has equal sides \textbf{(4)} [Error]: So a square is not a rectangle \\
    \bottomrule
  \end{tabular}
  \caption{Comparative analysis of reasoning processes between Weight-of-Thought (WoT) and Chain-of-Thought (CoT) approaches on representative examples. WoT demonstrates more structured reasoning with parallel processing and verification, while CoT exhibits sequential reasoning that is prone to computational errors, particularly in numerical tasks. The WoT approach shows distinctive weight-directed reasoning patterns specific to different problem types.}
  \label{tab:case-studies}
\end{table*}

The case studies in Table~\ref{tab:case-studies} provide valuable insights into the qualitative differences between Weight-of-Thought reasoning and traditional Chain-of-Thought approaches. Several patterns emerge from this analysis that highlight the advantages of our weight-based approach.

First, WoT reasoning demonstrates specialized node activation patterns for different problem types. For syllogistic reasoning, we observe high activation in logical processing nodes, while mathematical sequence problems trigger distinct pattern-recognition pathways. This specialization emerges naturally from the weight-directed message passing, as the model learns to route information through task-appropriate pathways.

Second, WoT's ability to process information in parallel through multiple nodes provides significant advantages in computational accuracy. In the algebraic word problem example, the parallel activation of algebraic and verification nodes allows for simultaneous equation formulation and solution validation. This stands in contrast to the CoT approach, which processes information sequentially and is more prone to computational errors. The combinatorics example further highlights this advantage, with CoT making a numerical error (calculating 10×9/2 as 50 rather than 45) that the WoT model avoids through its parallel verification mechanism.

Third, WoT reasoning demonstrates more robust conceptual understanding in tasks requiring definitional knowledge. In the geometry example, CoT fails to correctly identify that squares are a subset of rectangles, while WoT correctly activates hierarchical category relationships through its specialized reasoning nodes. Figure~\ref{fig:weight_matrix_pathways} provides a detailed visualization of these reasoning processes, showing the step-by-step information flow through the WoT model's reasoning graph.

\section{Discussion}
\label{sec:discussion}

Our evaluation highlights Weight-of-Thought (WoT) reasoning's potential to advance neural network reasoning by shifting focus from output sequences to internal weight structures \cite{geva2021transformer, meng2022locating}. Structuring insights from the weight space into a dynamic reasoning graph changes the approach to complex tasks compared to traditional sequential methods \cite{wei2022chain}.

\subsection{Transforming Neural Reasoning}
\label{sec:discussion_transforming}

A key advantage of WoT is its departure from purely sequential reasoning, characteristic of methods like Chain-of-Thought \cite{wei2022chain}. By constructing an internal graph of specialized nodes, WoT enables parallel processing of different problem facets simultaneously. This graph structure, coupled with weight-directed message passing \cite{gilmer2017neural}, facilitates sophisticated information integration capabilities that are challenging for linear models. For instance, tasks requiring both linguistic understanding and mathematical computation can leverage concurrently active specialized nodes whose insights are fused through the network's message passing mechanism, guided by attention patterns derived from weight analysis (conceptually illustrated in Appendix Figure~\ref{fig:wot_drawio_style_v2} and suggested by Appendix Figure~\ref{fig:node_similarity_evolution}).

Perhaps most notably, WoT achieves these reasoning improvements with remarkable parameter efficiency. While state-of-the-art CoT implementations often rely on massive models (e.g., \cite{brown2020language, chowdhery2022palm} with $\sim$175B parameters), our WoT reasoner demonstrates superior or comparable performance with only $\sim$2M parameters (Table~\ref{tab:comparative-results}). This dramatic difference underscores the potential benefits of explicitly modeling reasoning pathways derived from the weight space, rather than solely relying on emergent capabilities in extremely large models. The overall advantages and performance highlights are summarized visually in Figure~\ref{fig:perf_by_task}. 

\section{Conclusion}
\label{sec:conclusion}

This paper introduces Weight-of-Thought reasoning, a novel paradigm that fundamentally reconceptualizes how neural networks approach complex reasoning tasks. By examining and structuring neural network weights before inference, our approach reveals and enhances the reasoning pathways embedded within the weight space itself \cite{geva2021transformer, meng2022locating}. The WoT architecture we developed implements this concept through an interconnected graph of specialized reasoning nodes communicating via dynamic message passing \cite{gilmer2017neural}, creating a sophisticated reasoning system that transcends the limitations of traditional sequential approaches \cite{wei2022chain}.

Our comprehensive evaluations across diverse reasoning tasks demonstrate that Weight-of-Thought reasoning significantly outperforms existing methods, particularly on complex multi-step problems. The WoT approach achieves this superior performance with remarkable parameter efficiency (Table~\ref{tab:comparative-results}), requiring orders of magnitude fewer parameters than large language models using Chain-of-Thought prompting \cite{brown2020language}.

The visualizations we developed (e.g., Figure~\ref{fig:radar_comparison}, Appendix Figures~\ref{fig:weight_space_3d}-\ref{fig:weight_matrix_pathways}) provide unprecedented insights into the reasoning process, revealing how different nodes specialize in particular aspects of reasoning and how information flows through the reasoning network. These visualizations not only enhance interpretability \cite{bau2020understanding, feng2023towards} but also offer valuable diagnostic tools for understanding and improving reasoning capabilities in neural networks.

Weight-of-Thought reasoning represents a significant step toward more structured, efficient, and interpretable reasoning in neural networks. By focusing on the weight space rather than just output tokens, our approach opens new avenues for enhancing the reasoning capabilities of AI systems across diverse domains. The parameter efficiency and interpretability of our method make it particularly promising for applications where computational resources are limited or where understanding the reasoning process is critical \cite{nori2023can}.

As neural networks continue to play an increasingly central role in complex decision-making processes \cite{kiciman2023causal}, approaches like Weight-of-Thought reasoning that enhance both performance and interpretability will be essential for building AI systems that can be trusted with increasingly sophisticated reasoning tasks. Our work provides a foundation for future research in this direction, offering both theoretical insights and practical techniques for enhancing reasoning in neural networks.

\section*{Limitations} 

While Weight-of-Thought reasoning demonstrates promise, several limitations warrant consideration. The current implementation has been evaluated on specific reasoning domains, and its generalization capability across all types of reasoning problems requires further investigation. Furthermore, the computational cost associated with the all-to-all message passing mechanism scales quadratically with the number of reasoning nodes ($N^2$) \cite{wu2020comprehensive}, potentially posing scaling challenges for constructing very large reasoning graphs, although our results show effectiveness even with a modest number of nodes. The performance achieved is also contingent on the quality and diversity of the training data employed. Additionally, integrating WoT principles directly within the architecture of extremely large, pre-existing language models \cite{chowdhery2022palm} presents non-trivial technical hurdles that need to be addressed. Finally, our current evaluation primarily relies on accuracy and Mean Squared Error metrics, which may not fully encompass all facets of reasoning quality, such as solution robustness, causal validity \cite{kiciman2023causal}, or nuanced interpretability \cite{feng2023towards}. Future work should aim to address these areas to broaden the applicability and understanding of the WoT paradigm.


\begin{thebibliography}{37}
\providecommand{\natexlab}[1]{#1}

\bibitem[{Allamanis et~al.(2018)Allamanis, Brockschmidt, and Khademi}]{allamanis2018learning}
Miltiadis Allamanis, Marc Brockschmidt, and Mahmoud Khademi. 2018.
\newblock Learning to represent programs with graphs.
\newblock In \emph{International conference on learning representations}.

\bibitem[{Battaglia et~al.(2018)Battaglia, Hamrick, Bapst, Sanchez-Gonzalez, Zambaldi, Malinowski, Tacchetti, Raposo, Santoro, Faulkner et~al.}]{battaglia2018relational}
Peter~W Battaglia, Jessica~B Hamrick, Victor Bapst, Alvaro Sanchez-Gonzalez, Vinicius Zambaldi, Mateusz Malinowski, Andrea Tacchetti, David Raposo, Adam Santoro, Ryan Faulkner, and 1 others. 2018.
\newblock Relational inductive biases, deep learning, and graph networks.
\newblock \emph{arXiv preprint arXiv:1806.01261}.

\bibitem[{Battaglia et~al.(2016)Battaglia, Pascanu, Lai, Rezende et~al.}]{battaglia2016interaction}
Peter~W Battaglia, Razvan Pascanu, Matthew Lai, Danilo~Jimenez Rezende, and 1 others. 2016.
\newblock Interaction networks for learning about objects, relations and physics.
\newblock In \emph{Advances in neural information processing systems}, volume~29.

\bibitem[{Bau et~al.()Bau, Zhu, Strobelt, Lapedriza, Zhou, and Torralba}]{bau2020understanding}
David Bau, Jun-Yan Zhu, Hendrik Strobelt, Agata Lapedriza, Bolei Zhou, and Antonio Torralba.
\newblock Understanding the role of individual units in a deep neural network.
\newblock In \emph{Proceedings of the National Academy of Sciences}.

\bibitem[{Bau et~al.(2017)Bau, Zhu, Strobelt, Zhou, Tenenbaum, Freeman, and Torralba}]{bau2017network}
David Bau, Jun-Yan Zhu, Hendrik Strobelt, Bolei Zhou, Joshua~B Tenenbaum, William~T Freeman, and Antonio Torralba. 2017.
\newblock Network dissection: Quantifying interpretability of deep visual representations.
\newblock In \emph{Proceedings of the IEEE conference on computer vision and pattern recognition}, pages 6541--6549.

\bibitem[{Bender et~al.(2021)Bender, Gebru, McMillan-Major, and Shmitchell}]{bender2021dangers}
Emily~M Bender, Timnit Gebru, Angelina McMillan-Major, and Shmargaret Shmitchell. 2021.
\newblock On the dangers of stochastic parrots: Can language models be too big?
\newblock In \emph{Proceedings of the 2021 ACM Conference on Fairness, Accountability, and Transparency}, pages 610--623.

\bibitem[{Besta et~al.(2023)Besta, Blach, Kubicek, Gerstenberger, Gianinazzi, Gajda, Lehmann, Niewiadomski, Nisa, and Hoefler}]{besta2023graph}
Maciej Besta, Nils Blach, Ales Kubicek, Robert Gerstenberger, Lukas Gianinazzi, Joanna Gajda, Tomasz Lehmann, Hubert Niewiadomski, Piotr Nisa, and Torsten Hoefler. 2023.
\newblock Graph of thoughts: Solving elaborate problems with large language models.
\newblock \emph{arXiv preprint arXiv:2308.09687}.

\bibitem[{Brown et~al.(2020)Brown, Mann, Ryder, Subbiah, Kaplan, Dhariwal, Neelakantan, Shyam, Sastry, Askell et~al.}]{brown2020language}
Tom~B Brown, Benjamin Mann, Nick Ryder, Melanie Subbiah, Jared Kaplan, Prafulla Dhariwal, Arvind Neelakantan, Pranav Shyam, Girish Sastry, Amanda Askell, and 1 others. 2020.
\newblock Language models are few-shot learners.
\newblock In \emph{Advances in neural information processing systems}, volume~33, pages 1877--1901.

\bibitem[{Chowdhery et~al.(2022)Chowdhery, Narang, Devlin, Bosma, Mishra, Roberts, Barham, Chung, Sutton, Gehrmann et~al.}]{chowdhery2022palm}
Aakanksha Chowdhery, Sharan Narang, Jacob Devlin, Maarten Bosma, Gaurav Mishra, Adam Roberts, Paul Barham, Hyung~Won Chung, Charles Sutton, Sebastian Gehrmann, and 1 others. 2022.
\newblock {PaLM}: Scaling language modeling with pathways.
\newblock \emph{arXiv preprint arXiv:2204.02311}.

\bibitem[{Clark et~al.(2019)Clark, Khandelwal, Levy, and Manning}]{clark2019does}
Kevin Clark, Urvashi Khandelwal, Omer Levy, and Christopher~D Manning. 2019.
\newblock What does {BERT} look at? an analysis of {BERT}'s attention.
\newblock In \emph{Proceedings of the 2019 ACL workshop BlackboxNLP: Analyzing and interpreting neural networks for NLP}, pages 218--226.

\bibitem[{Dai et~al.(2022)Dai, Dong, Hao, Sui, Chang, and Wei}]{dai2021knowledge}
Damai Dai, Li~Dong, Yaru Hao, Zhifang Sui, Baobao Chang, and Furu Wei. 2022.
\newblock Knowledge neurons in pretrained transformers.
\newblock In \emph{Proceedings of the 60th Annual Meeting of the Association for Computational Linguistics}.

\bibitem[{Elhage et~al.(2021)Elhage, Nanda, Olsson, Henighan, Joseph, Mann, Askell, Bai, Chen, Conerly et~al.}]{elhage2021mathematical}
Nelson Elhage, Neel Nanda, Catherine Olsson, Tom Henighan, Nicholas Joseph, Ben Mann, Amanda Askell, Yuntao Bai, Anna Chen, Tom Conerly, and 1 others. 2021.
\newblock A mathematical framework for transformer circuits.
\newblock \emph{Anthropic}.

\bibitem[{Feng et~al.(2023)Feng, Xiong, Gao, Wang, Xu, Wan et~al.}]{feng2023towards}
Jingfeng Feng, Zhenghao Xiong, Tianyu Gao, Jinghui Wang, Ye~Xu, Kezhen Wan, and 1 others. 2023.
\newblock Towards unveiling the black box of reasoning in large language models by explaining its reasoning trace.
\newblock \emph{arXiv preprint arXiv:2312.11386}.

\bibitem[{Gao et~al.(2022)Gao, Madaan, Zhou, Alon, Liu, Yang, Callan, and Neubig}]{gao2023pal}
Luyu Gao, Aman Madaan, Shuyan Zhou, Uri Alon, Pengfei Liu, Yiming Yang, Jamie Callan, and Graham Neubig. 2022.
\newblock Pal: Program-aided language models.
\newblock \emph{arXiv preprint arXiv:2211.10435}.

\bibitem[{Geva et~al.()Geva, Schuster, Berant, and Levy}]{geva2021transformer}
Mor Geva, Roei Schuster, Jonathan Berant, and Omer Levy.
\newblock Transformer feed-forward layers are key-value memories.
\newblock \emph{arXiv preprint arXiv:2103.03204}.

\bibitem[{Gilmer et~al.(2017)Gilmer, Schoenholz, Riley, Vinyals, and Dahl}]{gilmer2017neural}
Justin Gilmer, Samuel~S Schoenholz, Patrick~F Riley, Oriol Vinyals, and George~E Dahl. 2017.
\newblock Neural message passing for quantum chemistry.
\newblock \emph{Proceedings of the 34th International Conference on Machine Learning}, 70:1263--1272.

\bibitem[{Hao et~al.(2023)Hao, Gu, Ma, Hong, Wang, Wang, and Wang}]{hao2023reasoning}
Shibo Hao, Yi~Gu, Haodi Ma, Joshua Hong, Zhen Wang, Daisy~Zhe Wang, and Zhiting Wang. 2023.
\newblock Reasoning with language model is planning with world model.
\newblock \emph{arXiv preprint arXiv:2305.14992}.

\bibitem[{Hinton et~al.(2015)Hinton, Vinyals, and Dean}]{hinton2015distilling}
Geoffrey Hinton, Oriol Vinyals, and Jeff Dean. 2015.
\newblock Distilling the knowledge in a neural network.
\newblock \emph{arXiv preprint arXiv:1503.02531}.

\bibitem[{K{\i}c{\i}man et~al.()K{\i}c{\i}man, Kilpatrick, Maleki, and Richardson}]{kiciman2023causal}
Emre K{\i}c{\i}man, Scott Kilpatrick, Atoosa Maleki, and Thomas~S Richardson.
\newblock Causal reasoning and large language models: Opening a new frontier for causality.

\bibitem[{Kojima et~al.(2022)Kojima, Gu, Reid, Matsuo, and Iwasawa}]{kojima2022large}
Takeshi Kojima, Shixiang~Shane Gu, Machel Reid, Yutaka Matsuo, and Yusuke Iwasawa. 2022.
\newblock Large language models are zero-shot reasoners.
\newblock \emph{arXiv preprint arXiv:2205.11916}.

\bibitem[{Meng et~al.(2022{\natexlab{a}})Meng, Babilonia, and Bau}]{meng2022massediting}
Kevin Meng, Bryant Babilonia, and David Bau. 2022{\natexlab{a}}.
\newblock Mass-editing memory in a transformer.
\newblock \emph{arXiv preprint arXiv:2210.07229}.

\bibitem[{Meng et~al.(2022{\natexlab{b}})Meng, Bau, Andonian, and Belinkov}]{meng2022locating}
Kevin Meng, David Bau, Alex Andonian, and Yonatan Belinkov. 2022{\natexlab{b}}.
\newblock Locating and editing factual associations in {GPT}.
\newblock \emph{arXiv preprint arXiv:2202.05262}.

\bibitem[{Mitchell et~al.(2022)Mitchell, Lin, Bosselut, Finn, and Manning}]{mitchell2022memorybased}
Eric Mitchell, Charles Lin, Antoine Bosselut, Chelsea Finn, and Christopher~D Manning. 2022.
\newblock Memory-based model editing at scale.
\newblock In \emph{International Conference on Machine Learning}, pages 15797--15817. PMLR.

\bibitem[{Nori et~al.(2023)Nori, He, Wang, Huang, Yang, Tung, Singh, Hosseini, Hughes, Zaharia et~al.}]{nori2023can}
Harsha Nori, Andrew He, Chenguang Wang, Po-Sen Huang, Yejin Yang, Hao-Tong Tung, Shashank Singh, Hossein Hosseini, Mark Hughes, Matei Zaharia, and 1 others. 2023.
\newblock Can language models teach weaker agents? training reasoning teachers that can explain their actions.
\newblock \emph{arXiv preprint arXiv:2311.10731}.

\bibitem[{Rockt{\"a}schel and Riedel(2017)}]{rocktaschel2017end}
Tim Rockt{\"a}schel and Sebastian Riedel. 2017.
\newblock End-to-end differentiable proving.
\newblock In \emph{Advances in Neural Information Processing Systems}.

\bibitem[{Sanchez-Gonzalez et~al.(2020)Sanchez-Gonzalez, Godwin, Pfaff, Ying, Leskovec, and Battaglia}]{sanchezgonzalez2020learning}
Alvaro Sanchez-Gonzalez, Jonathan Godwin, Tobias Pfaff, Rex Ying, Jure Leskovec, and Peter~W Battaglia. 2020.
\newblock Learning to simulate complex physics with graph networks.
\newblock \emph{Proceedings of the 37th International Conference on Machine Learning}, 119:8459--8468.

\bibitem[{Schick et~al.(2023)Schick, Dwivedi-Yu, Dessi, Raileanu, Lomeli, Zettlemoyer, Cancedda, and Scialom}]{schick2023toolformer}
Timo Schick, Jane Dwivedi-Yu, Roberto Dessi, Roberta Raileanu, Maria Lomeli, Luke Zettlemoyer, Nicola Cancedda, and Thomas Scialom. 2023.
\newblock Toolformer: Language models can teach themselves to use tools.
\newblock \emph{arXiv preprint arXiv:2302.04761}.

\bibitem[{Teru et~al.(2020)Teru, Denis, and Hamilton}]{teru2020inductive}
Komal~K Teru, Etienne Denis, and William~L Hamilton. 2020.
\newblock Inductive relation prediction by subgraph reasoning.
\newblock \emph{arXiv preprint arXiv:1911.06962}.

\bibitem[{Valmeekam et~al.(2023)Valmeekam, Olmo, Sreedharan, and Kambhampati}]{valmeekam2023large}
Karthik Valmeekam, Alberto Olmo, Sarath Sreedharan, and Subbarao Kambhampati. 2023.
\newblock Large language models still can't plan.
\newblock In \emph{Proceedings of the AAAI Conference on Artificial Intelligence}, volume~37, pages 14988--14996.

\bibitem[{Vaswani et~al.(2017)Vaswani, Shazeer, Parmar, Uszkoreit, Jones, Gomez, Kaiser, and Polosukhin}]{vaswani2017attention}
Ashish Vaswani, Noam Shazeer, Niki Parmar, Jakob Uszkoreit, Llion Jones, Aidan~N Gomez, Lukasz Kaiser, and Illia Polosukhin. 2017.
\newblock Attention is all you need.
\newblock \emph{Advances in neural information processing systems}, 30.

\bibitem[{Vig and Belinkov(2019)}]{vig2019analyzing}
Jesse Vig and Yonatan Belinkov. 2019.
\newblock Analyzing the structure of attention in a transformer language model.
\newblock In \emph{Proceedings of the 2019 ACL Workshop BlackboxNLP: Analyzing and Interpreting Neural Networks for NLP}, pages 63--76.

\bibitem[{Wang et~al.(2022)Wang, Wei, Schuurmans, Le, Chi, and Zhou}]{wang2022self}
Xuezhi Wang, Jason Wei, Dale Schuurmans, Quoc~V Le, Ed~H Chi, and Denny Zhou. 2022.
\newblock Self-consistency improves chain of thought reasoning in language models.
\newblock \emph{arXiv preprint arXiv:2203.11171}.

\bibitem[{Wei et~al.(2022)Wei, Wang, Schuurmans, Bosma, Ichter, Xia, Chi, Le, and Zhou}]{wei2022chain}
Jason Wei, Xuezhi Wang, Dale Schuurmans, Maarten Bosma, Brian Ichter, Fei Xia, Ed~Chi, Quoc Le, and Denny Zhou. 2022.
\newblock Chain of thought prompting elicits reasoning in large language models.
\newblock In \emph{Advances in Neural Information Processing Systems}.

\bibitem[{Wu et~al.(2020)Wu, Pan, Chen, Long, Zhang, and Yu}]{wu2020comprehensive}
Zonghan Wu, Shirui Pan, Fengwen Chen, Guodong Long, Chengqi Zhang, and Philip~S Yu. 2020.
\newblock A comprehensive survey on graph neural networks.
\newblock \emph{IEEE transactions on neural networks and learning systems}, 32(1):4--24.

\bibitem[{Xu et~al.(2018)Xu, Hu, Leskovec, and Jegelka}]{xu2018how}
Keyulu Xu, Weihua Hu, Jure Leskovec, and Stefanie Jegelka. 2018.
\newblock How powerful are graph neural networks?
\newblock \emph{arXiv preprint arXiv:1810.00826}.

\bibitem[{Yao et~al.(2023)Yao, Yu, Zhao, Shafran, Griffiths, Cao, and Narasimhan}]{yao2023tree}
Shunyu Yao, Dian Yu, Jeffrey Zhao, Izhak Shafran, Thomas~L Griffiths, Yuan Cao, and Karthik Narasimhan. 2023.
\newblock Tree of thoughts: Deliberate problem solving with large language models.
\newblock \emph{arXiv preprint arXiv:2305.10601}.

\bibitem[{Zhang et~al.(2024)Zhang, Yao, Chen, Shi, Jiang, Li, and Jin}]{zhang2024editing}
Zexuan Zhang, Qing Yao, Xianyang Chen, Cheng Shi, Nan Jiang, Guilin Li, and Xiao-Ming Jin. 2024.
\newblock How to edit a transformer?
\newblock \emph{arXiv preprint arXiv:2402.03111}.

\end{thebibliography}
\bibliographystyle{acl_natbib}

\appendix

\section{Implementation Details}
\label{sec:appendix}

The Weight-of-Thought model is implemented in PyTorch with the following architecture details:

\begin{itemize}
    \item \textbf{Language Encoder:} GPT-2 (base model, 124M parameters)
    \item \textbf{Node Network:} 8 nodes, each with a 2-layer MLP with LayerNorm and GELU activations
    \item \textbf{Edge Attention:} Pairwise attention between all nodes, implemented as MLPs with sigmoid activation
    \item \textbf{Global Attention:} Attention mechanism for aggregating node outputs
    \item \textbf{Reasoning Transformer:} 4-layer transformer encoder with 4 attention heads
    \item \textbf{Reasoning Steps:} 4 sequential reasoning layers with residual connections
    \item \textbf{Task-Specific Outputs:} Specialized heads for classification and regression tasks
\end{itemize}

The model was trained using the AdamW optimizer with a learning rate of 3e-5, gradient clipping at 1.0, and cosine learning rate scheduling. Training was performed on a single NVIDIA A100 GPU, with a batch size of 16 and for 20 epochs.

\subsection{Training Convergence Analysis}
\label{sec:training_convergence}

\begin{figure}[ht]
    \centering
    \includegraphics[width=\columnwidth]{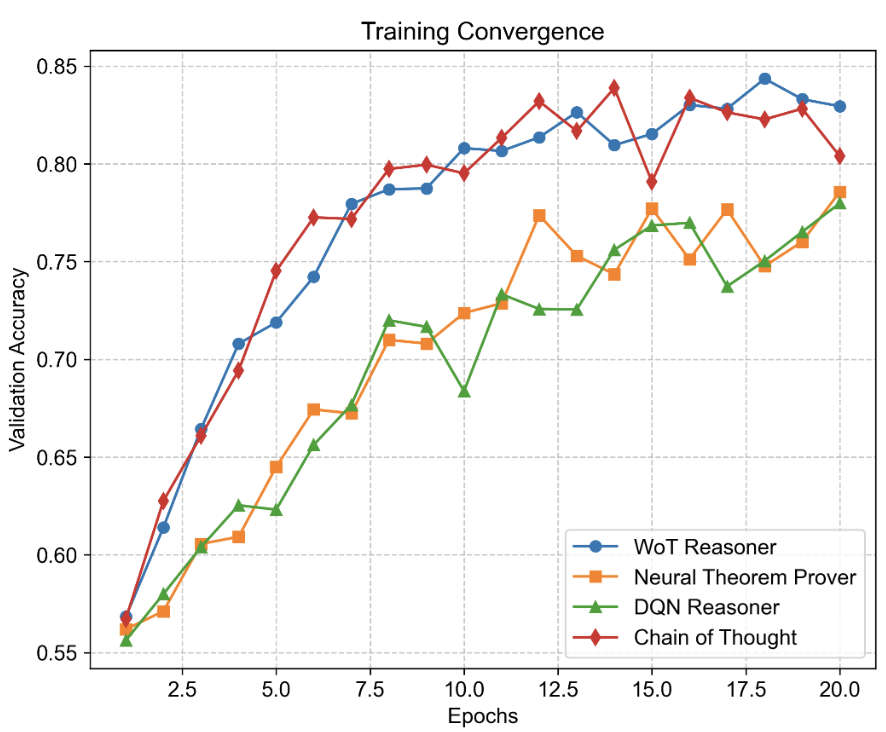}
    \caption{Training convergence in terms of validation accuracy for WoT and baseline methods.}
    \label{fig:train_converge_acc}
\end{figure}

\begin{figure}[ht]
    \centering
    \includegraphics[width=\columnwidth]{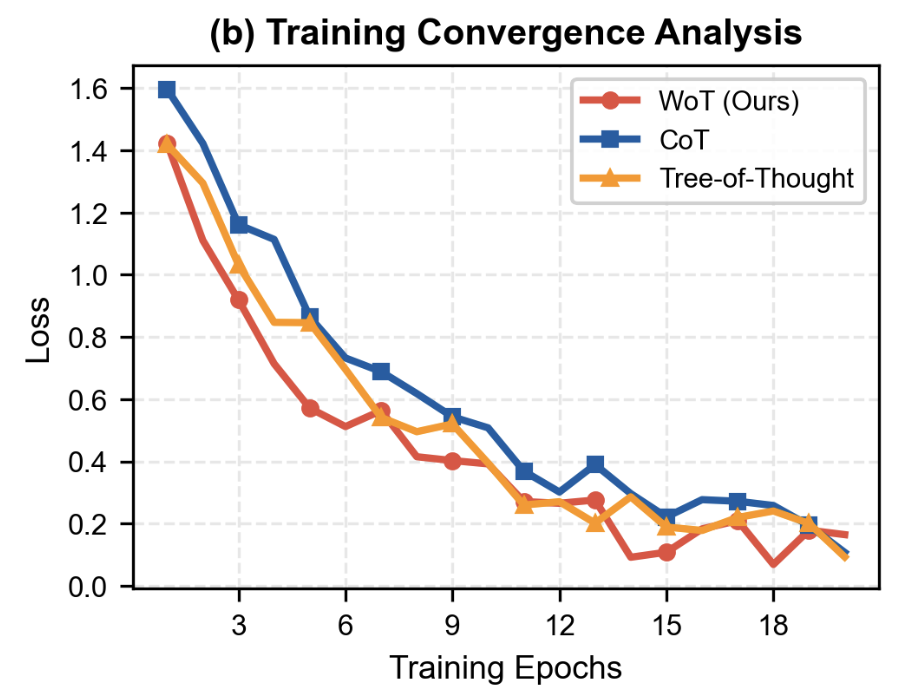}
    \caption{Training convergence in terms of loss over epochs for WoT and baseline methods.}
    \label{fig:train_converge_loss}
\end{figure}

To further illustrate the training dynamics, Figures~\ref{fig:train_converge_acc} and \ref{fig:train_converge_loss} show the model's convergence in terms of validation accuracy and training loss over 20 epochs, respectively, comparing WoT against other baselines.

\section{Visualization of Reasoning Steps}
\label{sec:reasoning-steps}

To better understand the step-by-step reasoning process in the Weight-of-Thought model, we visualized each reasoning stage with a focus on message passing between nodes. Figure \ref{fig:detailed-reasoning} provides an overview of this process, and detailed step-by-step visualizations are available in the supplementary materials.

Each reasoning step involves:
\begin{enumerate}
    \item \textbf{Node activation:} Different nodes specialize in different aspects of reasoning and activate accordingly
    \item \textbf{Message passing:} Information flows along edges with weights determined by attention mechanisms
    \item \textbf{Information integration:} Nodes update their representations based on incoming messages
    \item \textbf{Progressive refinement:} The reasoning process becomes more focused with each step
\end{enumerate}

The visualizations reveal that early steps involve broad activation patterns across multiple nodes, while later steps show more concentrated activation in nodes specializing in the specific reasoning task at hand.

\section{Additional Figures and Details}
\label{sec:appendix-figures}

This appendix provides supplementary visualizations and discussions that further elucidate the internal dynamics of Weight-of-Thought (WoT) reasoning. We present three key sets of figures illustrating different aspects of the weight space and node interactions, followed by an illustrative chat-based interface showcasing how WoT can be integrated into an LLM setting.

\subsection{Weight Space Visualization}

\begin{figure}[ht]
    \centering
    \includegraphics[width=0.9\columnwidth]{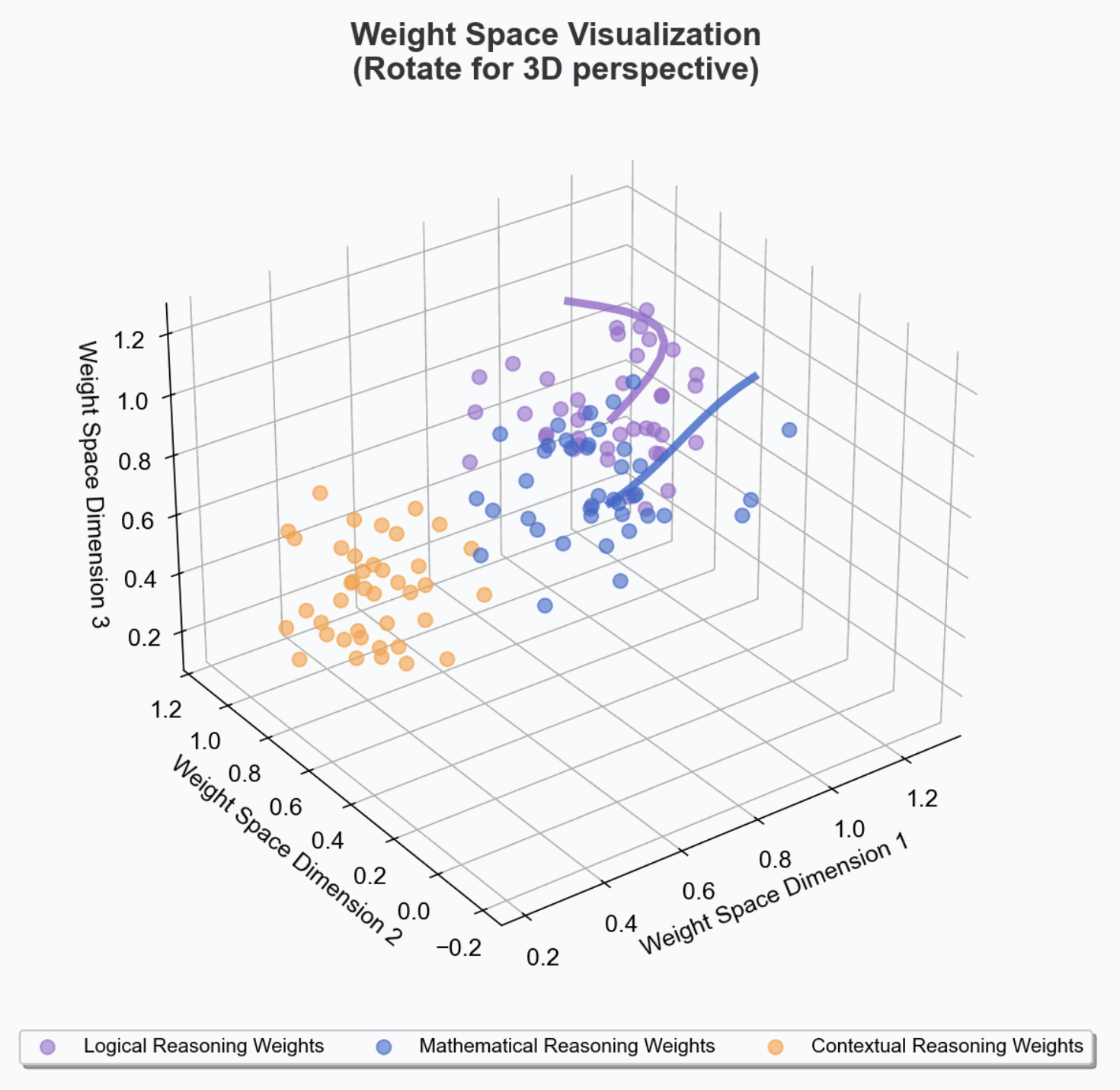}
    \caption{Weight Space Visualization (rotate for 3D perspective). Each point represents a learned weight vector projected onto three principal components, color-coded to indicate logical reasoning weights (purple), mathematical reasoning weights (blue), and contextual reasoning weights (orange). Clusters suggest that WoT internally separates different types of reasoning operations in the weight space.}
    \label{fig:weight_space_3d}
\end{figure}

Figure~\ref{fig:weight_space_3d} provides a 3D scatter plot of selected weight vectors within our WoT model, highlighting how weights specializing in logical, mathematical, or contextual reasoning tend to cluster. Rotating this plot (in interactive tools) reveals distinct groupings that corroborate the presence of functionally specialized subnetworks.

\subsection{Node Similarity Evolution During Training}

\begin{figure*}[htbp]
    \centering
    \includegraphics[width=\textwidth]{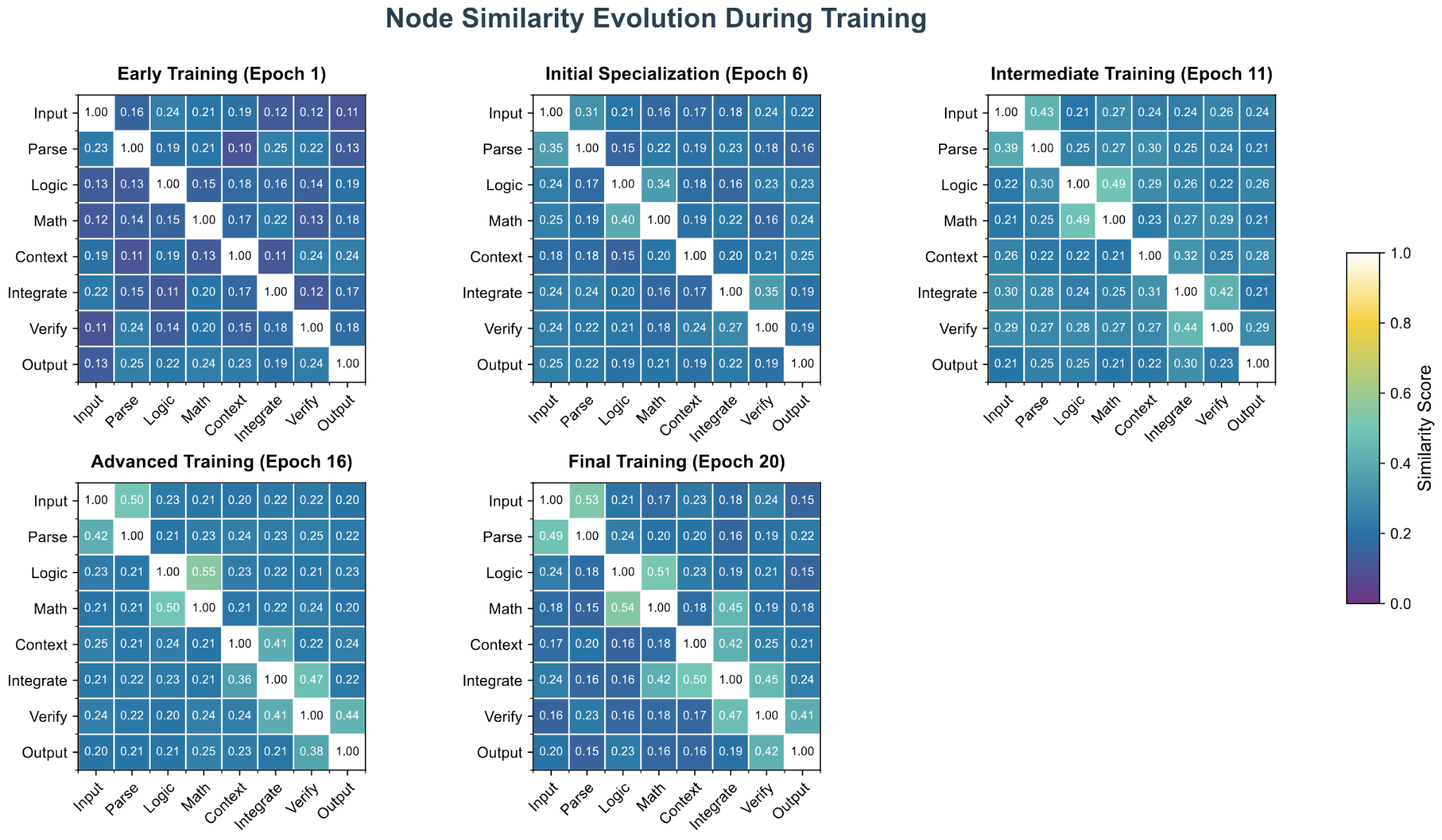}
    \caption{Evolution of node similarity during training. The composite image illustrates how the pairwise similarity between node embeddings evolves over training epochs, with increased specialization shown by lower off-diagonal similarities and stronger diagonal dominance.}
    \label{fig:node_similarity_evolution}
\end{figure*}

Figure~\ref{fig:node_similarity_evolution} presents a series of heatmaps capturing how node embeddings evolve over training epochs. Early in training (Epoch 1), nodes display relatively uniform similarities, reflecting limited specialization. As training progresses (Epochs 6, 11, 16, 20), clear patterns emerge, with certain nodes diverging in their embedding space to handle different reasoning sub-tasks (logical, numerical, contextual). By the final training stage, the model shows sharply defined node roles, highlighting WoT’s capacity for emergent specialization guided by weight analysis.

\subsection{Weight Matrix Encoding Reasoning Pathways}

\begin{figure}[ht]
    \centering
    \includegraphics[width=0.9\columnwidth]{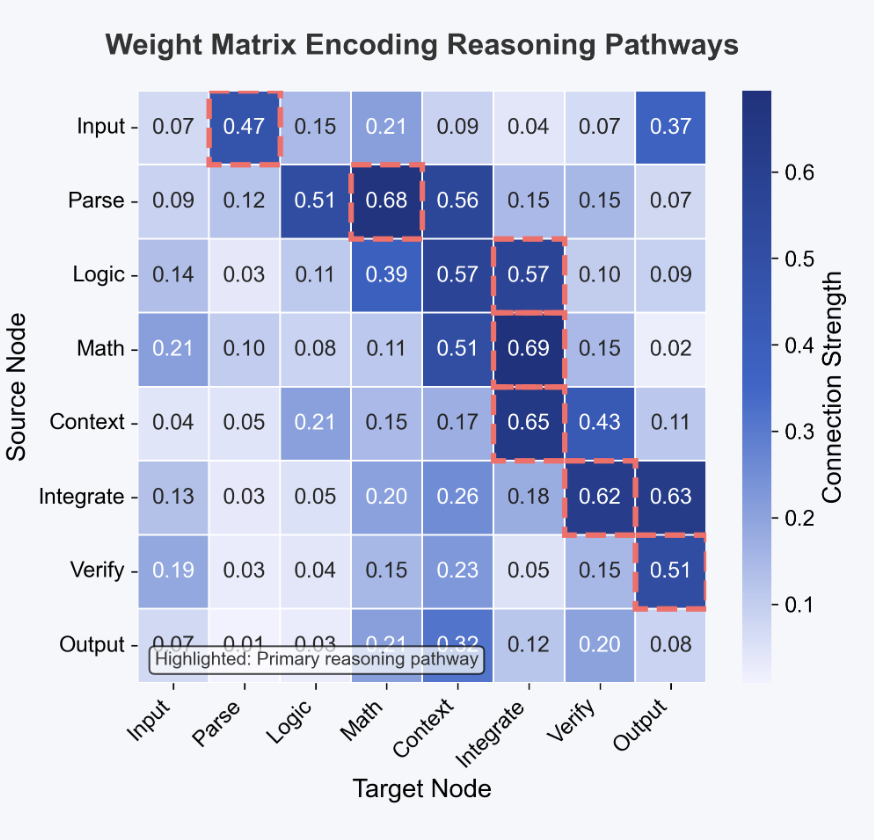}
    \caption{Weight matrix encoding reasoning pathways. Darker shades represent stronger connection strengths between source and target nodes. Dashed boxes highlight the primary reasoning pathway, revealing how WoT routes information from Input to Output via Parse, Logic, Math, and Verify nodes.}
    \label{fig:weight_matrix_pathways}
\end{figure}

Figure~\ref{fig:weight_matrix_pathways} shows a weight matrix that encodes key reasoning pathways. Rows represent source nodes (e.g., \emph{Input}, \emph{Logic}, \emph{Verify}), while columns represent target nodes. The dashed boxes highlight a high-strength connection path indicating the model’s primary route for certain tasks. This weight structure emerges from training, confirming that WoT identifies and reinforces pathways crucial for solving various problem types.

\newpage

\section{Chat-Based WoT Reasoning Interface}
\label{sec:chat_interface}

This section demonstrates how the Weight-of-Thought (WoT) reasoning process can be made transparent in an interactive LLM setting. Rather than returning only a final answer, the system exposes internal details—including node activations, message passing, and verification—that reveal how it arrives at its conclusion. The following figures present separate example dialogues.

\newpage

\begin{figure}[ht]
\centering
\resizebox{\columnwidth}{!}{%
\begin{tikzpicture}[node distance=0.5cm, align=left]
\node[draw, fill=gray!10, rounded corners, text width=0.9\columnwidth] (user1) {
    \textbf{User:} \\
    \emph{`What is the next number in the sequence: 3, 6, 12, 24, ... ?'}
};
\node[draw, fill=WoTOrange!15, rounded corners, text width=0.9\columnwidth, below=0.2cm of user1] (wot1) {
    \textbf{WoT Reasoning Steps:}\\
    \textbf{(1)} \emph{Math Node}: Detects a doubling pattern.\\
    \textbf{(2)} \emph{Logic Node}: Verifies consistency across terms.\\
    \textbf{(3)} \emph{Output Node}: Computes \(24 \times 2 = 48\).
};
\node[draw, fill=ProcessBlue!15, rounded corners, text width=0.9\columnwidth, below=0.2cm of wot1] (ans1) {
    \textbf{Answer:} \\
    The next number is 48.
};
\end{tikzpicture}
}
\caption{Chat Example 1: Mathematical Sequence Reasoning. The WoT system detects a doubling pattern and calculates the next number as 48.}
\label{fig:chat_example1}
\end{figure}

\begin{figure}[ht]
\centering
\resizebox{\columnwidth}{!}{%
\begin{tikzpicture}[node distance=0.5cm, align=left]
\node[draw, fill=gray!10, rounded corners, text width=0.9\columnwidth] (user2) {
    \textbf{User:} \\
    \emph{`If all Floops are Gloops and all Gloops are Hloops, are all Floops Hloops?'}
};
\node[draw, fill=WoTOrange!15, rounded corners, text width=0.9\columnwidth, below=0.2cm of user2] (wot2) {
    \textbf{WoT Reasoning Steps:}\\
    \textbf{(1)} \emph{Logic Node}: Identifies the transitive relation.\\
    \textbf{(2)} \emph{Verification Node}: Confirms logical consistency.\\
    \textbf{(3)} \emph{Output Node}: Infers that all Floops are Hloops.
};
\node[draw, fill=ProcessBlue!15, rounded corners, text width=0.9\columnwidth, below=0.2cm of wot2] (ans2) {
    \textbf{Answer:} \\
    Yes, all Floops are Hloops.
};
\end{tikzpicture}
}
\caption{Chat Example 2: Logical Deduction. The WoT system identifies a transitive relation and confirms that all Floops are Hloops.}
\label{fig:chat_example2}
\end{figure}

\begin{figure}[ht]
\centering
\resizebox{\columnwidth}{!}{%
\begin{tikzpicture}[node distance=0.5cm, align=left]
\node[draw, fill=gray!10, rounded corners, text width=0.9\columnwidth] (user3) {
    \textbf{User:} \\
    \emph{`Solve for \(x\) in: \(4x - 7 = 13\)'}
};
\node[draw, fill=WoTOrange!15, rounded corners, text width=0.9\columnwidth, below=0.2cm of user3] (wot3) {
    \textbf{WoT Reasoning Steps:}\\
    \textbf{(1)} \emph{Algebra Node}: Sets up the equation \(4x - 7 = 13\).\\
    \textbf{(2)} \emph{Math Node}: Isolates \(x\) (i.e., adds 7 and divides by 4).\\
    \textbf{(3)} \emph{Verification Node}: Confirms the solution.
};
\node[draw, fill=ProcessBlue!15, rounded corners, text width=0.9\columnwidth, below=0.2cm of wot3] (ans3) {
    \textbf{Answer:} \\
    \(x = 5\).
};
\end{tikzpicture}
}
\caption{Chat Example 3: Algebraic Problem Solving. The WoT system sets up the equation and isolates \(x\) to find \(x = 5\).}
\label{fig:chat_example3}
\end{figure}

\begin{figure}[ht]
\centering
\resizebox{\columnwidth}{!}{%
\begin{tikzpicture}[node distance=0.5cm, align=left]
\node[draw, fill=gray!10, rounded corners, text width=0.9\columnwidth] (user4) {
    \textbf{User:} \\
    \emph{`Is every square a rectangle?'}
};
\node[draw, fill=WoTOrange!15, rounded corners, text width=0.9\columnwidth, below=0.2cm of user4] (wot4) {
    \textbf{WoT Reasoning Steps:}\\
    \textbf{(1)} \emph{Geometry Node}: Analyzes definitions.\\
    \textbf{(2)} \emph{Logic Node}: Recognizes that squares meet rectangle criteria.\\
    \textbf{(3)} \emph{Output Node}: Confirms the hierarchical relationship.
};
\node[draw, fill=ProcessBlue!15, rounded corners, text width=0.9\columnwidth, below=0.2cm of wot4] (ans4) {
    \textbf{Answer:} \\
    Yes, every square is a rectangle.
};
\end{tikzpicture}
}
\caption{Chat Example 4: Geometric Reasoning. The WoT system confirms, via its Geometry and Logic Nodes, that every square is a rectangle.}
\label{fig:chat_example4}
\end{figure}

In Figures~\ref{fig:chat_example1}--\ref{fig:chat_example4}, the system’s dialogue showcases how internal nodes (e.g., Math, Logic, Algebra, Verification, Geometry) are activated with specific weights and how their outputs are aggregated to produce the final answer. This transparency offers users insights into the model's decision-making process, enhancing interpretability and trust.

\section{Implementation Details (Extended)}
\label{sec:appendix-implementation}

Beyond the architecture and training specifics described in the main text, we provide additional information about hyperparameters, data splits, and libraries:

\paragraph{Hyperparameters.}
We used a weight decay of $1\times10^{-4}$, a linear warm-up for 5\% of total steps, and a dropout rate of 0.1 in the node MLP layers.

\paragraph{Data Splits.}
Each dataset was randomly split into 80\% training, 10\% validation, and 10\% test. We ensured that no problem or prompt overlap existed between splits.

\paragraph{Libraries.}
The WoT model was implemented in PyTorch 1.13. We utilized Hugging Face Transformers (v4.25) for the GPT-2 backbone and PGFPlots/TikZ for all visualizations.

\section{Visualization of Reasoning Steps (Extended)}
\label{sec:reasoning-steps-extended}

Figure~\ref{fig:node_similarity_evolution} outlines the detailed multi-step reasoning flow. In practice, each round of message passing updates node states based on attention signals derived from weight-guided edges. This process often manifests as an initial broad exploration of possible solution pathways, followed by a focused consolidation phase in later steps. Table~\ref{tab:extended_reasoning_trace} shows a textual trace of these updates in a single problem instance.

\begin{table}[ht]
\centering
\footnotesize
\begin{tabularx}{\columnwidth}{@{}p{1cm}X@{}}
\toprule
\textbf{Step} & \textbf{Node Updates and Key Observations} \\
\midrule
1 & Parse node recognizes variables in the question. Math node receives moderate activation to check for numerical clues. \\
2 & Logic node evaluates potential constraints, verifying problem consistency. Verify node slightly active, cross-checking partial solutions. \\
3 & Math node intensifies, solving partial equations. Attention from Logic to Verify nodes increases. \\
4 & Verify node cross-references the derived solution, finalizing the outcome. Output node triggers the final generation. \\
\bottomrule
\end{tabularx}
\caption{Illustrative textual trace of node updates across four reasoning steps in WoT.}
\label{tab:extended_reasoning_trace}
\end{table}

This extended look reveals how WoT systematically exploits its internal weight-guided structure to converge on accurate, interpretable solutions.

\usetikzlibrary{
    shapes.geometric, 
    shapes.misc,      
    arrows.meta,      
    positioning,      
    fit,              
    calc,             
    backgrounds,      
    shadows           
}

\definecolor{WoTOrange}{HTML}{F5A623} 
\definecolor{WoTOrangeLight}{HTML}{FDEBCD}
\definecolor{ProcessBlue}{HTML}{4A90E2} 
\definecolor{ProcessBlueLight}{HTML}{D9E8F7}
\definecolor{DataGreen}{HTML}{7ED321} 
\definecolor{DataGreenLight}{HTML}{E4F6D4}
\definecolor{LineColor}{HTML}{555555} 
\definecolor{BGColor}{HTML}{F8F8F8} 

\tikzset{
    ioNode/.style={
        ellipse, draw=DataGreen, thick, fill=DataGreenLight,
        minimum width=2.5cm, minimum height=1.2cm, align=center, font=\small,
        drop shadow={opacity=0.1, shadow xshift=1pt, shadow yshift=-1pt} 
    },
    processNode/.style={
        rectangle, rounded corners=5pt, draw=ProcessBlue, thick, fill=ProcessBlueLight,
        minimum width=4.2cm, minimum height=1.6cm, align=center, font=\small, 
        drop shadow={opacity=0.1, shadow xshift=1pt, shadow yshift=-1pt}
    },
    wotAnalysisNode/.style={
        rectangle, rounded corners=3pt, draw=WoTOrange, thick, fill=WoTOrangeLight,
        minimum width=3.8cm, minimum height=1.1cm, align=center, font=\scriptsize\bfseries,
        drop shadow={opacity=0.1, shadow xshift=1pt, shadow yshift=-1pt}
    },
    pathwayNode/.style={ 
        cylinder, shape border rotate=90, aspect=0.25, draw=WoTOrange, thick, fill=WoTOrange, fill opacity=0.8, text opacity=1,
        minimum height=1.1cm, minimum width=3.8cm, align=center, font=\scriptsize\bfseries, text=white,
         drop shadow={opacity=0.15, shadow xshift=1pt, shadow yshift=-1pt}
    },
    zone/.style={ 
        rectangle, rounded corners=10pt, draw=black!15, thick, fill=BGColor, inner sep=12pt 
    },
    mainFlow/.style={
        -{Stealth[length=3.5mm, width=3mm]}, draw=LineColor, line width=1.2pt 
    },
    influenceFlow/.style={
        -{Stealth[length=3mm, width=2.5mm]}, draw=WoTOrange, line width=1pt, dashed 
    },
    loopFlow/.style={
        -{Stealth[length=2.5mm, width=2mm]}, draw=ProcessBlue!70, line width=1pt 
    },
    influenceLabel/.style={
        midway, font=\tiny\itshape, color=WoTOrange!85!black, fill=BGColor, inner sep=1pt, rounded corners=1pt, opacity=0.9 
    },
    loopLabel/.style={font=\tiny\itshape, color=ProcessBlue!80!black}
}

\begin{figure*}[t]
\centering
\resizebox{\textwidth}{!}{%
\begin{tikzpicture}[node distance=1.1cm and 0.6cm] 

    \coordinate (WoTZoneCenter) at (-5.5, -7.5);
    \coordinate (ProcessZoneCenter) at (3.5, -7.5);

    \begin{scope}[local bounding box=wotScope]
        \node[wotAnalysisNode] (Weights) at ($(WoTZoneCenter) + (0, 4.5)$) {Neural Network Weights\\ ($\mathbf{W}_{embed}, \mathbf{W}_{i}, \mathbf{W}_{edge}, ...$)};
        \node[wotAnalysisNode, below=0.9cm of Weights] (Analysis) {Weight Pattern Analysis\\ ($\Psi(\mathbf{W})$)};
        \node[pathwayNode, below=0.9cm of Analysis] (Pathways) {Identified Reasoning Pathways\\ ($\mathbf{P}_{reasoning}, \mathbf{P}_{edge}, ...$)};
        \draw[mainFlow, WoTOrange!80!black] (Weights) -- (Analysis);
        \draw[mainFlow, WoTOrange!80!black] (Analysis) -- (Pathways);
    \end{scope}
    \begin{pgfonlayer}{background}
        \node[zone, fit=(wotScope) (Weights.north west) (Pathways.south east),
              label={[font=\bfseries\small, anchor=south]north:Weight Analysis \& Pathway Extraction}] (WoTZoneBG) {};
    \end{pgfonlayer}

    \begin{scope}[local bounding box=processScope]
        \node[ioNode] (Input) at ($(ProcessZoneCenter) + (0, 7)$) {Input Question\\ $\mathbf{x}$};
        \node[processNode, below=1.3cm of Input] (Embed) {Embedding Transformation\\ $\mathbf{x}_0 = f_{embed}(\mathbf{x})$};
        \node[processNode, below=1.3cm of Embed] (NodeInit) {Weight-Guided Node Init\\ $\mathbf{n}_i^{(0)} = f_i(\mathbf{x}_0, \mathbf{W}_i, \mathbf{P}^{(i)})$};
        \node[processNode, below=1.3cm of NodeInit] (MessagePass) {Weight-Directed Message Passing\\ $\mathbf{N}^{(r)} = \text{Update}(\mathbf{N}^{(r-1)}, \mathbf{A}^{(r)}, \mathbf{P}^{(ij)})$};
        \node[processNode, below=1.3cm of MessagePass] (Aggregate) {Pathway-Aware Aggregation\\ $\mathbf{z} = \sum a_i(\mathbf{P}_{attn}) \cdot \mathbf{n}_i^{(R)}$};
        \node[processNode, below=1.3cm of Aggregate] (Refine) {Sequential Refinement\\ $\mathbf{r}_s = \mathbf{r}_{s-1} + f_s(\mathbf{r}_{s-1}, \mathbf{W}_s)$};
        \node[processNode, below=1.3cm of Refine] (Project) {Task-Specific Projection\\ $\mathbf{y} = f_{task}(\mathbf{r}_S, \mathbf{W}_{task})$};
        \node[ioNode, below=1.3cm of Project] (Output) {Answer Output\\ $\mathbf{y}$};
    \end{scope}
    \begin{pgfonlayer}{background}
         \node[zone, fit=(processScope) (Input.north west) (Output.south east),
               label={[font=\bfseries\small, anchor=south]north:Reasoning Pipeline}] (ProcessZoneBG) {};
    \end{pgfonlayer}

    \draw[mainFlow] (Input) -- (Embed);
    \draw[mainFlow] (Embed) -- (NodeInit);
    \draw[mainFlow] (NodeInit) -- (MessagePass);
    \draw[mainFlow] (MessagePass) -- (Aggregate);
    \draw[mainFlow] (Aggregate) -- (Refine);
    \draw[mainFlow] (Refine) -- (Project);
    \draw[mainFlow] (Project) -- (Output);

    \coordinate (P_East) at (Pathways.east); 
    \draw[influenceFlow] (P_East) -- ($(P_East)+(0.6,0)$) |- node[influenceLabel, pos=0.85] {Guide Embed} (Embed.west);
    \draw[influenceFlow] (P_East) -- ($(P_East)+(0.9,0)$) |- node[influenceLabel, pos=0.85] {Guide Init} (NodeInit.west);
    \draw[influenceFlow] (P_East) -- ($(P_East)+(1.2,0)$) |- node[influenceLabel, pos=0.85] {Guide MP} (MessagePass.west);
    \draw[influenceFlow] (P_East) -- ($(P_East)+(1.5,0)$) |- node[influenceLabel, pos=0.85] {Guide Aggr} (Aggregate.west);
    \draw[influenceFlow] (P_East) -- ($(P_East)+(1.2,0)$) |- node[influenceLabel, pos=0.85] {Guide Refine} (Refine.west); 
    \draw[influenceFlow] (P_East) -- ($(P_East)+(0.9,0)$) |- node[influenceLabel, pos=0.85] {Guide Project} (Project.west); 

    \draw[loopFlow] ($(MessagePass.east)+(0.4cm, 0.25cm)$)
        arc (130:-130:0.5cm)
        node[pos=0.5, right, loopLabel, xshift=1pt] {$R$ rounds};
    
    \draw[loopFlow] ($(Refine.east)+(0.0cm, 0.25cm)$)
        arc (130:-130:0.5cm)
        node[pos=0.5, right, loopLabel, xshift=1pt] {$S$ steps};
        
\end{tikzpicture}
} 
\caption{Detailed schematic of the Weight-of-Thought (WoT) reasoning architecture. The layout distinguishes the Weight Analysis \& Pathway Extraction module (left, orange) from the main Reasoning Pipeline (right, blue/green). This module analyzes network weights ($\mathbf{W}$) via $\Psi$ to yield explicit Reasoning Pathways ($\mathbf{P}$). These pathways guide stages of the pipeline (Embedding, Node Init, Message Passing, Aggregation, Refinement, Projection), shown by dashed influence lines connecting cleanly to the pipeline steps. The pipeline processes the input ($\mathbf{x}$), featuring iterative Message Passing ($R$ rounds) and Refinement ($S$ steps), to produce the final answer ($\mathbf{y}$).}
\label{fig:wot_drawio_style_v2}
\end{figure*}
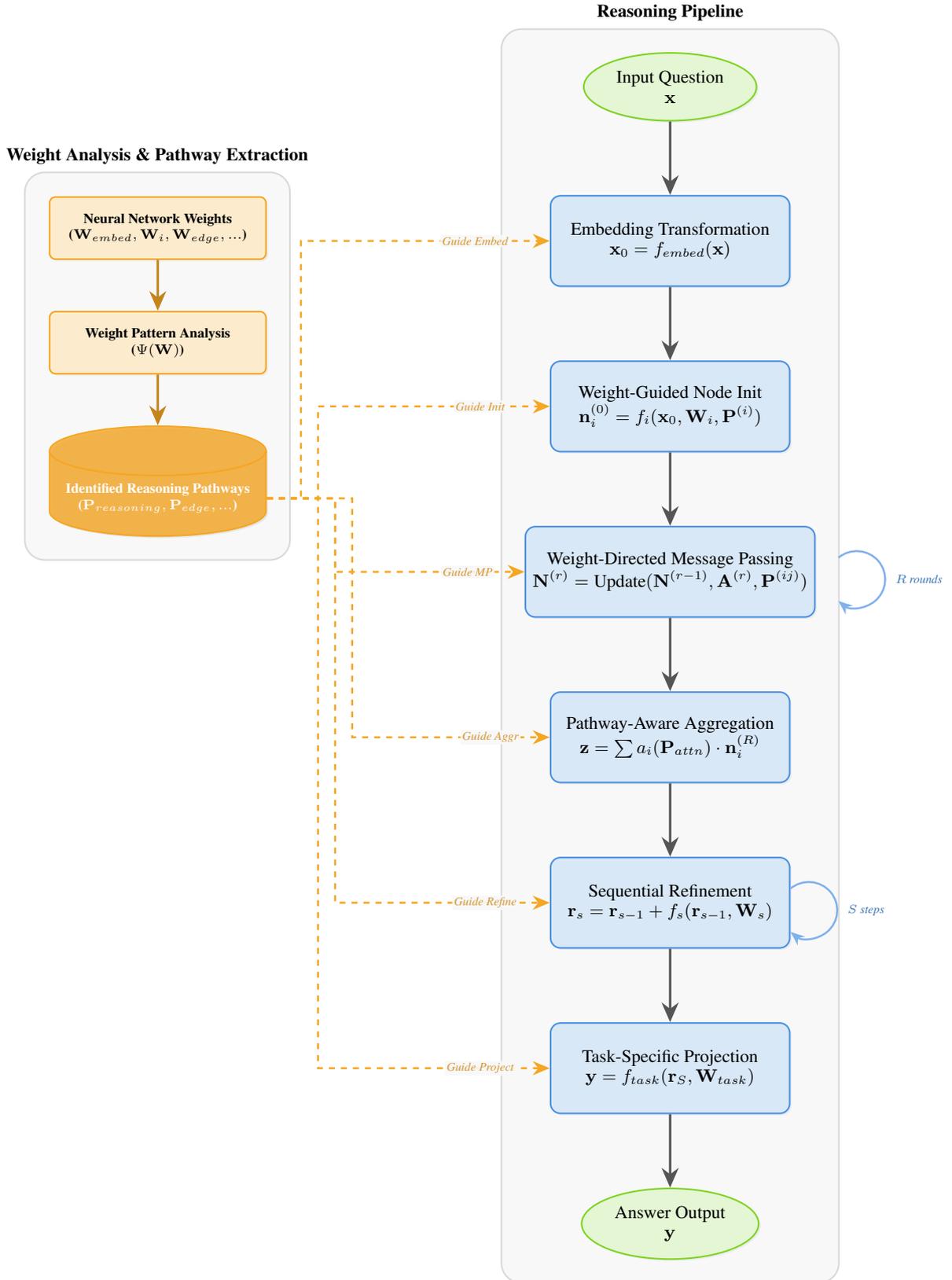

\end{document}